\documentclass[10pt,twocolumn,letterpaper]{article}

\usepackage{iccv}
\usepackage{times}
\usepackage{epsfig}
\usepackage{graphicx}
\usepackage{amsmath}
\usepackage{amssymb}

\usepackage{booktabs}
\usepackage{multirow}

\usepackage[pagebackref=true,breaklinks=true,letterpaper=true,colorlinks,bookmarks=false]{hyperref}

\iccvfinalcopy % *** Uncomment this line for the final submission

\def\iccvPaperID{****} % *** Enter the ICCV Paper ID here

\ificcvfinal\fi

\begin{document}

%%%%%%%%% TITLE
\title{Real-World Video for Zoom Enhancement based on Spatio-Temporal Coupling}

% \author{Zhiling Guo\\
% The Hong Kong Polytechnic University\\
% Kowloon, Hong Kong, China\\
% {\tt\small zhiling.guo@polyu.edu.hk}
% \and
% Yinqiang Zheng\\
% The University of Tokyo\\
% Bunkyo City, Tokyo 113-0033, Japan\\
% {\tt\small yqzheng@ai.u-tokyo.ac.jp}
% \and
% Haoran Zhang\\
% Peking University\\
% Shenzhen, Guangdong, China\\
% {\tt\small h.zhang@pku.edu.cn}
% \and
% Xiaodan Shi\\
% The University of Tokyo\\
% Kashiwa 277-8568, Japan\\
% {\tt\small shixiaodan@csis.u-tokyo.ac.jp}
% \and
% Zekun Cai\\
% The University of Tokyo\\
% Kashiwa 277-8568, Japan\\
% {\tt\small caizekun@csis.u-tokyo.ac.jp}
% \and
% Ryosuke Shibasaki\\
% The University of Tokyo\\
% Kashiwa 277-8568, Japan\\
% {\tt\small shiba@csis.u-tokyo.ac.jp}
% \and
% Jinyue Yan\\
% The Hong Kong Polytechnic University\\
% Kowloon, Hong Kong, China\\
% {\tt\small j-jerry.yan@polyu.edu.hk}
% }

\author{
Zhiling Guo$^{\textit{1,2}}$,
Yinqiang Zheng$^{\textit{3}}$,
Haoran Zhang$^{\textit{4}}$,
Xiaodan Shi$^{\textit{2}}$,
Zekun Cai$^{\textit{2}}$, \\
Ryosuke Shibasaki$^{\textit{2}}$,
Jinyue Yan$^{\textit{1}}$
\\
\\
$^{\textit{1}}$Department of Building Environment and Energy Engineering, \\ The Hong Kong Polytechnic University, Kowloon, Hong Kong, China
\\ $^{\textit{2}}$Center for Spatial Information Science, The University of Tokyo, Kashiwa, Japan
\\ $^{\textit{3}}$Next Generation Artificial Intelligence Research Center, The University of Tokyo, Tokyo, Japan
\\ $^{\textit{4}}$School of Urban Planning and Design, Peking University, Shenzhen, China
}

% \address{$^{\textit{1}}$Department of Building Environment and Energy Engineering, The Hong Kong Polytechnic University, Kowloon, Hong Kong, China
% \\ $^{\textit{2}}$Center for Spatial Information Science, the University of Tokyo, Kashiwa, Japan
% \\ $^{\textit{3}}$Center for Spatial Information Science, the University of Tokyo, Kashiwa, Japan
% \\ $^{\textit{4}}$School of Urban Planning and Design, Peking University, No.2199 Lishui Road, Nanshan District, Shenzhen, Guangdong, 518055, China
% }

% \address{$^{\textit{1}}$Department of Building Environment and Energy Engineering, The Hong Kong Polytechnic University, Kowloon, Hong Kong, China
% \\ $^{\textit{2}}$Center for Spatial Information Science, the University of Tokyo, Kashiwa, Japan
% \\ $^{\textit{3}}$Center for Spatial Information Science, the University of Tokyo, Kashiwa, Japan
% \\ $^{\textit{4}}$School of Urban Planning and Design, Peking University, No.2199 Lishui Road, Nanshan District, Shenzhen, Guangdong, 518055, China
% }

\maketitle
\ificcvfinal\thispagestyle{empty}\fi

%%%%%%%%% ABSTRACT
\begin{abstract}
In recent years, single-frame image super-resolution (SR) has become more realistic by considering the zooming effect and using real-world short- and long-focus image pairs. In this paper, we further investigate the feasibility of applying realistic multi-frame clips to enhance zoom quality via spatio-temporal information coupling. Specifically, we first built a real-world video benchmark, VideoRAW, by a synchronized co-axis optical system. The dataset contains paired short-focus raw and long-focus sRGB videos of different dynamic scenes. Based on VideoRAW, we then presented a Spatio-Temporal Coupling Loss, termed as STCL. The proposed STCL is intended for better utilization of information from paired and adjacent frames to align and fuse features both temporally and spatially at the feature level. The outperformed experimental results obtained in different zoom scenarios demonstrate the superiority of integrating real-world video dataset and STCL into existing SR models for zoom quality enhancement, and reveal that the proposed method can serve as an advanced and viable tool for video zoom.
\end{abstract}

%%%%%%%%% BODY TEXT
\section{Introduction}

\begin{figure}[t]
\centering
\includegraphics[width=\columnwidth]{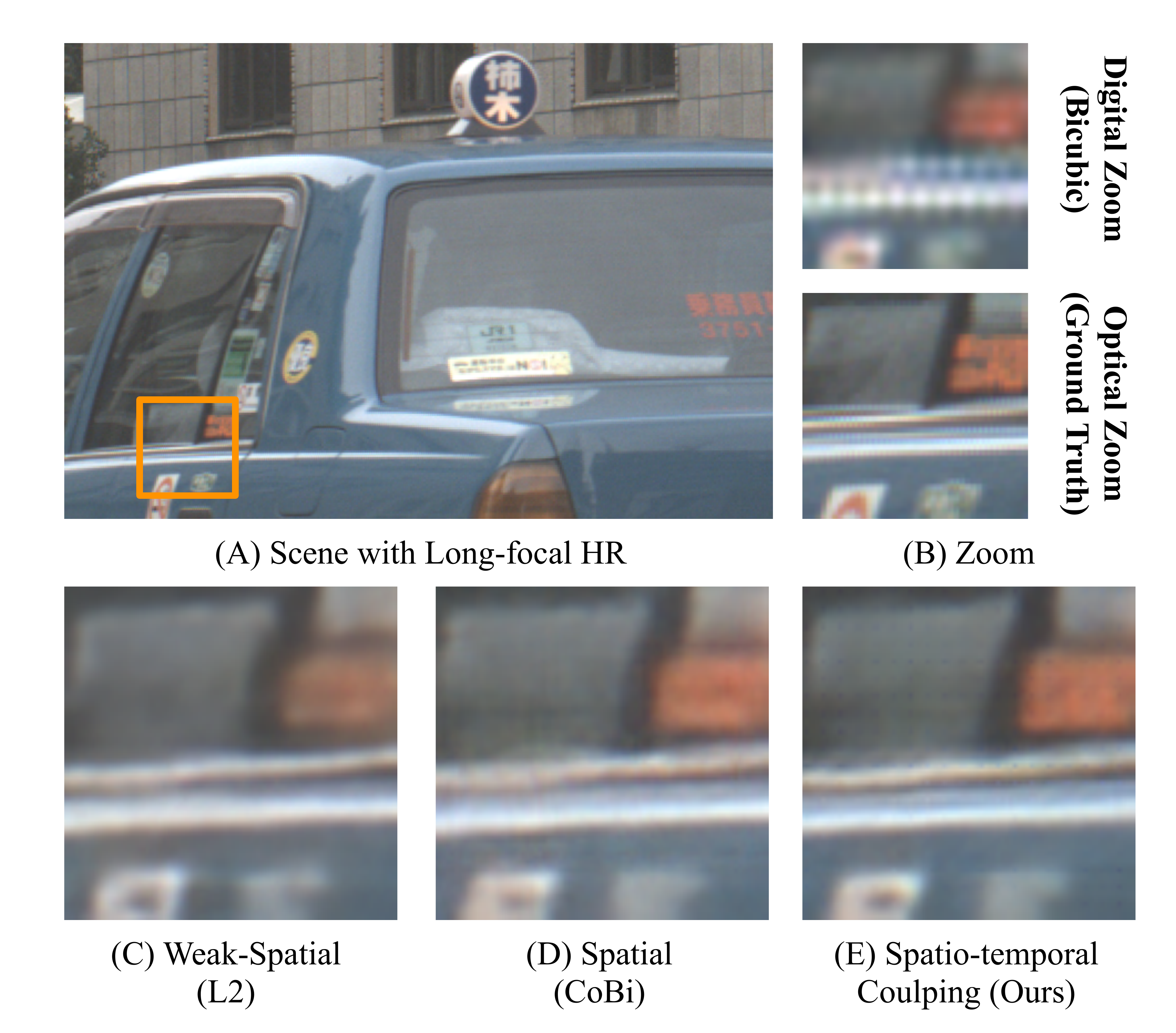}
\caption{Visual comparison among realistic LR\&HR pair and learning based digital zoom methods. $L_{2}$, CoBi, and Ours all trained by SRResNet architecture, here represent the methods based on weak-spatial, spatial, and spatio-temporal coupling constraint, respectively.}
\label{fig:teaser}
\end{figure}

Zoom functionality plays an important role nowadays owing to the increasing demand for more detailed contents of view in camera equipment. Instead of using the bulky and expensive optical lens, adopting digital zoom to increase the resolution has emerged as an alternative strategy. Regarding the digital zoom method, which accomplished by cropping a portion of an image down to a centered area and simply upsampling back up to the same aspect ratio as the original, as shown in Figure \ref{fig:teaser} B, it unavoidably results in the annoying quality problems in super-resolved image including noise, artifacts, loss of detail, unnaturalness, etc. \cite{sheikh2016methodology}. The generalization of high-quality contents based on digital zoom remains a formidable challenge.

% 超分的可以用来做，但zoom与现有超分在本质上的不同
Considering the zoom principles as well as the difficulties faced by digital zoom mentioned above, super-resolution (SR) techniques, which aim at increasing the image resolution while providing finer spatial details than those captured by the original acquisition sensors, can be adopted to boost digital zoom quality. SR has experienced significant improvements over the last few years thanks to deep learning methods and large-scale training datasets \cite{dong2015image,shi2016real,kim2016accurate,lai2017deep,ledig2017photo,wang2018esrgan,dai2019second,zhang2020deep}. Unfortunately, most existing methods that evaluated on simulated datasets are hard to be generalized in challenging real-world SR conditions, where the authentic degradations in low-resolution (LR) images are far more complicated \cite{cai2019ntire,kohler2018bridging,yang2014single}. Thus, introducing high-quality training datasets that contain real-world LR and high-resolution (HR) pairs to realistic SR is highly desired.

% %-------------------------------------
% 别人是如何做zoom的，以及出现的一些问题
% 受到了 Cobi的启发用raw加上特殊的loss结合深度学习来进行zoom，题到Cobi开创了用raw来做zoom的先河，RAW数据的优点，但是单帧RAW数据的具有局限性
% 1. 数据本身的问题
% 2. 不能进行视频的拓展
% 3. 尽量提出三点问题 （看考超分视频相对于超分图片的优点）

Recent studies \cite{zhang2019zoom,cai2019toward,chen2019camera,Joze_2020_CVPR_Workshops} have investigated the strategy of applying real sensor based single-frame datasets including raw data for digital zoom quality enhancement. Although remarkable improvement can be achieved, since some severe issues are inevitable in captured raw data pairs, including lens distortion, spatial misalignment, and color mismatching, the effectiveness of SR is heavily limited by single-frame based method in how much information can be reconstructed from limited and misaligned spatial features.

% %-------------------------------------
% 我们提出的方法，用多帧视频来进行实验，进一步提出我们具体的方法。
The successes of applying simulated multi-frame datasets in image restoration tasks such as video SR \cite{Xiang_2020_CVPR, Isobe_2020_CVPR, yi2019progressive, wang2019edvr, yang2018multi} and deblurring \cite{tao2018scale} inspire us to raise the possibility of adopting real sensor based multi-frame datasets in digital zoom. Considering the inter-frame spatio-temporal correlations, extracting and combining information from multiple frames would be a promising strategy to alleviate the intrinsic issues between realistic LR and HR pairs. Under the aforementioned assumption, in this paper, we proposed to utilize real sensor based video datasets in SR to achieve digital zoom enhancement. It remains the challenges in two aspects: (1) how to acquire high-quality real-world video pairs with different resolutions, and (2) how to effectively make full use of the multi-frame information for model training.  

% 我们的光学系统
To obtain a paired video dataset, we build a novel optical system that adopts a beam splitter to split the light from the same scene, and then capture the paired LR and HR videos by the equipped long- and short-focal length camera independently and simultaneously. The system can conveniently collect realistic raw videos as well as image datasets with different scale ratios by simply adjusting the equipped manual zoom lens. In this paper, we provide a benchmark, named as VideoRAW, for training and evaluating SR algorithms in practical applications. We define a scene captured at a long focal length as the HR ground truth, and the same one captured at a short focal length as its paired LR observation. In comparison to existing datasets \cite{kohler2018bridging,qu2016capturing,cai2019toward,zhang2019zoom,cai2019ntire, Joze_2020_CVPR_Workshops}, VideoRAW is the first large-scale video-based raw dataset used for real-world computational zoom. It enables the comparisons among different algorithms in both video and image zoom scenarios, and the diverse scenes contained inside make it more realistic and practical. 

% where ten raw video pairs captured from different scenes are included for training and evaluating zoom algorithms in practical applications
% Here, we provide a benchmark dataset named VideoRAW, where ten raw video pairs captured from different scenes are included for training and evaluating zoom algorithms in practical applications.
% Here, we define a scene captured at a long focal length as the HR ground truth, and the same one captured at a short focal length as its paired LR observation. 

% Alignment(LR gaussian constraint loss)
% Fusion()
% Merge 
% spatial and temporay

% 只能通过loss来解决，网络不能解决
According to VideoRAW, where the paired LR and HR images are not perfectly aligned while adjacent frames contain spatio-temporal correlations, we propose a novel loss framework, termed Spatio-Temporal Coupling Loss (STCL), to address the challenges in training for the challenging feature alignment and fusion. STCL draws inspiration from recent proposed contextual bilateral loss (CoBi) \cite{zhang2019zoom} in dealing with unaligned features in paired single-frame LR and HR images. Different from CoBi, which only focuses on limited spatial pattern, STCL takes both spatial and temporal correlations of reference multi-frame clips into account, and performs realistic SR enhancement in a coupled manner at the feature level. Specifically, regarding spatial aspect, STCL aligns the location of HR and input LR in a lower scale with coarse constraint while fusing the features from the paired HR frame into SR in a higher scale. In perspective of temporary aspect, STCL convolves the features in adjacent frames and then takes them as supplementary cues to help compensate the SR quality.  

% %-------------------------------------
% We introduce a fundamentally different loss framework  in this paper. We propose a novel loss that achieve high performance zoom in four steps. 
% %-------------------------------------

% 综述我们要做的事情，在怎样的场景下测试的，如何评价以及取得的效果
Finally, the proposed VideoRAW and STCL are adopted to SR for digital zoom quality enhancement. During training, raw sensor data, which taken with a shorter focal length, are served as LR input to fully exploit the information from raw, as well as to avoid the artifacts occurred in demosaicing preprocessing \cite{gharbi2016deep,zhang2019zoom,chen2018learning}. To evaluate our approach, we integrate the proposed method into different existing deep learning based SR architectures \cite{ledig2017photo,lim2017enhanced,yamanaka2017fast,vu2018fast}. The experimental results show that our method could outperform others in both construction accuracy \cite{hore2010image} and perceptual quality \cite{zhang2018unreasonable,johnson2016perceptual} among all scenarios, which reveal the generalizability and the effeteness of applying real-world video datasets and spatio-temporal coupling method in realistic SR.

The main contributions of this study are three-fold:
\begin{itemize}
  \item We demonstrate the feasibility of introducing spatio-temporal coupling for zoom quality enhancement, which is achieved by adopting real-world zoom video datasets and a novel loss framework.
  \item We design a co-axial optical system to obtain paired short- and long-focal length videos from different scenes, and will publicly release a valuable real sensor based raw/sRGB video benchmark, VideoRAW.
  \item We present a loss framework named STCL in dealing with realistic SR based on VideoRAW and spatio-temporal coupling. 
\end{itemize}

It should be noted that our paper is the first realistic video-based solution for learning-based digital zoom quality enhancement.

\section{Related Work}
% 10.26
% To our best knowledge, in learning-based digital zoom quality enhancement task, there is no existing research concentrates on video datasets. Instead, a few of the recent studies adopting sing-frame datasets.
\subsection{Super-Resolution for Digital Zoom}
The past few years have witnessed great success in applying deep learning to enhance the SR quality \cite{ledig2017photo, lim2017enhanced, yamanaka2017fast, vu2018fast, dong2015image, shi2016real, kim2016accurate, lai2017deep, ledig2017photo, wang2018esrgan, dai2019second, Isobe_2020_CVPR, Xiang_2020_CVPR}. However, most existing SR methods are typically evaluated with simulated datasets, learning digital zoom in practical scenarios by these methods would be less effective and results in a significant deterioration on the SR performance.

Lately, the realistic single-frame datasets based methods are proposed by a few of the studies for zoom quality enhancement. Cai \etal \cite{cai2019toward} presented a new single-frame benchmark and a Laplacian pyramid based kernel predication network (LP-KPN) to handle the real-world scenes. Chen \etal \cite{chen2019camera} investigated the feasibility by alleviating the intrinsic tradeoff between resolution and field-of-view from the perspective of camera lenses. Zhang \etal \cite{zhang2019zoom} adopted a contextual bilateral loss (CoBi) to deal with the misalignment issue between paired LH and HR images, and used a high-bit raw dataset, SR-RAW, to improve the input data quality. The benefits of using raw as input over RGB is proven in \cite{zhang2019zoom} as well. Joze \etal \cite{Joze_2020_CVPR_Workshops} released a realistic dataset named ImagePairs for image SR and image quality enhancement. Besides, NTIRE 2019 \cite{cai2019ntire} organized a challenge to facilitate the development of real image SR. The provided image pairs in the challenge are registered beforehand by a sophisticated image registration algorithm, thus high-quality SR results can be generated by using carefully designed deep learning architectures\cite{cheng2019encoder,kwak2019fractal,zhang2018image}.
Considering that very accurate subpixel registration is difficult in realistic raw data, as well as the misaligned feature would undermine the feature extraction capability of deep learning model, instead of using single-frame pairs, to adopt temporal correlation in adjacent multi-frame clips would be a promising strategy to enhance the limited spatial information. However, to our best knowledge, there exists no work for learning-based zoom by multi-frame pairs, and the closest one is \cite{wronski2019handheld}, which supplants the need for demosaicing in a camera pipeline by merging a burst of raw images. In this paper, we address the challenge of introducing both spatial and temporal information via real-world video dataset for learning-based zoom quality enhancement.

\subsection{Spatio-Temporal Coupling}
%引入时间信息的方法容易被想到，如何耦合变得困难
% 时间信息的方法容易被想到，如何耦合变得困难
%引入时空想关心，但是运用网络在zoom上不能从本质上来解决耦合的问题
The inter-frame spatial and temporal information has been exploited by many recent simulated data based video SR studies \cite{wang2020deep, isobe2020video, Isobe_2020_CVPR, wang2019edvr, yi2019progressive, caballero2017real, xue2019video, tao2017detail, sajjadi2018frame}. The methods can be grouped into time-based, space-based, and spatio-temporal based. A representative study of time-based way is presented by Shi \etal \cite{xingjian2015convolutional}, which formulated a convolutional long short-term memory (ConvLSTM) architecture to reserve information from the last frame. Regarding space-based way, the aim of which is trying to merge temporal information in a parallel manner, studies such as VSRnet \cite{kappeler2016video} and DUFVSR \cite{jo2018deep} achieved the goal by direct fusion architecture and 3D convolutional neural networks (CNN), respectively. Besides, for spatio-temporal way, Yi \etal \cite{yi2019progressive} proposed a novel progressive fusion network by incorporating a series of progressive fusion residual blocks (PFRBs), and Caballero \etal \cite{caballero2017real} adopted spatio-temporal networks and motion compensation, and Wang \cite{wang2019edvr} introduced the enhanced deformable convolutional networks, while Wang \cite{wang2020deep} applied optical flow estimation. 

However, due to the spatial misalignment issues in realistic video pairs, the aforementioned methods which mainly focus on network architecture optimization have not been able to effectively couple spatio-temporal information via pixel-wise loss functions. Instead of further optimizing the network architecture, we propose to investigate the feasibility of achieving spatio-temporal coupling in the perspective of loss function.

\section{Realistic Video Raw Dataset}
% 10.26 10.27
To enable training, we introduce a novel dataset, VideoRAW, which contains realistic video pairs of both LR and HR, taken with our co-axis optical imaging system. For data preprocessing, we align the captured video pairs one-by-one with geometric transformation.
\begin{figure}[ht]
\centering
\includegraphics[width=\columnwidth]{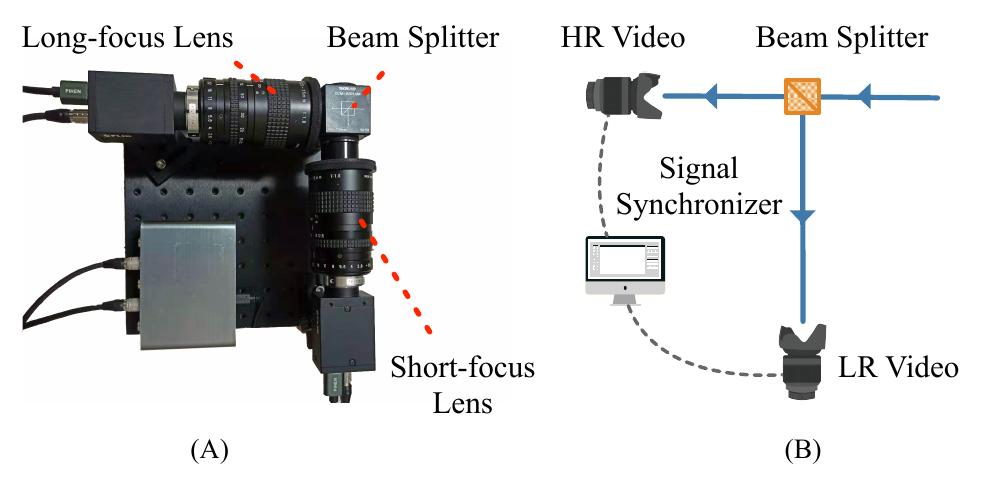}
\caption{Design of our optical system. (A) Real device equipped with identical manual zoom lenses in long- and short-focal lengths; and (B) Video capturing with an external signal generator to keep rigorous temporal synchronization.}
\label{fig:device}
\end{figure}

\subsection{Data Capturing}
As shown in Figure \ref{fig:device}, the optical system consists of one beam splitter, two global shutter cameras, two identical zoom lenses with different focal lengths, and a signal synchronizer to keep time synchronization. When capturing videos, the incoming light is first divided into two perpendicular lights by a 45$^{\circ}$ beam splitter, and then pass through a long and a short focal-length camera equipped with RGGB Bayer sensor, respectively.

Here, two FLIR GS3-U3-15S5C cameras and two RICOH FL-CC6Z1218-VG manual zoom lenses are adopted to collect 4X upscale ratio video pairs. The focal length is set to 18mm in the branch of LR videos with larger field-of-view (FoV), and it is set to 72mm in the branch of HR videos with smaller FoV. Although we focus on the investigation of 4X data, which is common in video SR, our capture system can be used to capture up to 6x paired data without modifications. 

For camera settings, we choose 15fps frame rate to enhance spatial-temporal correlation, 2mm shutter speed to avoid obvious blur on fast-moving objects, and a relative small aperture size to alleviate the influence of depth-of-field difference. With the proposed imaging system, 84 pairs in multiple scenes, each containing 200 frames with 1384 $\times$ 1032 resolution, are captured from different street spots. We take 16-bit LR raw and 8-bit HR sRGB as the input and ground truth for zoom learning, respectively. It should be noted that the camera has a 14-bit ADC and the affiliated Flycapture2 software converts the original 14-bit raw into 16-bit by linear scaling. The 8-bit HR sRGB images are generated by the in-built ISP of Flycapture2. 

\begin{figure}[ht]
\centering
\includegraphics[width=\columnwidth]{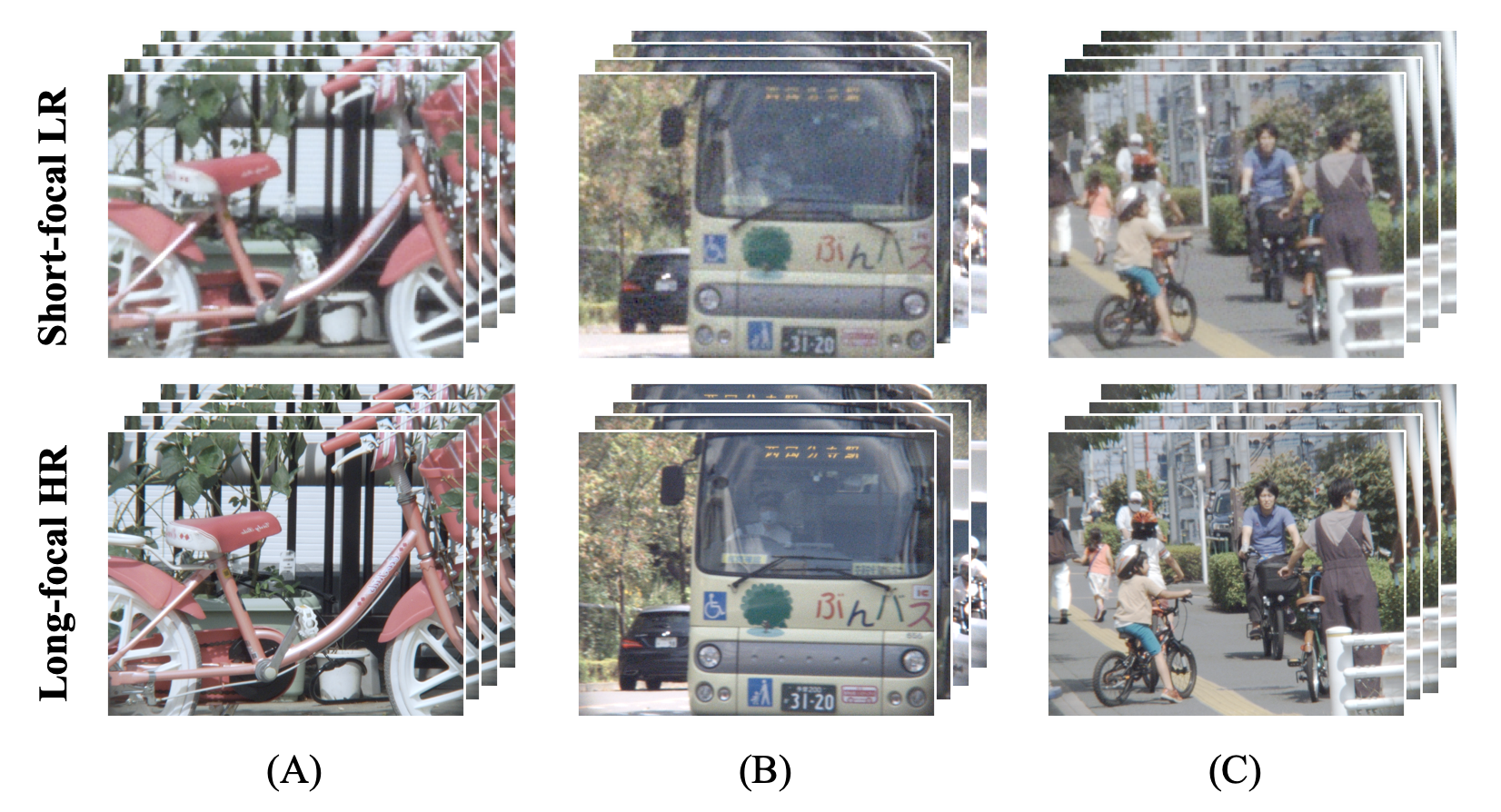}
\caption{A subset of the VideoRAW dataset. The top row includes consecutive LR frames after FoV matching, while the second row is the paired HR frames.}
\label{fig:scene_sample}
\end{figure}

\subsection{Data Prepossessing}
\begin{figure}[h!]
\centering
\includegraphics[width=\columnwidth]{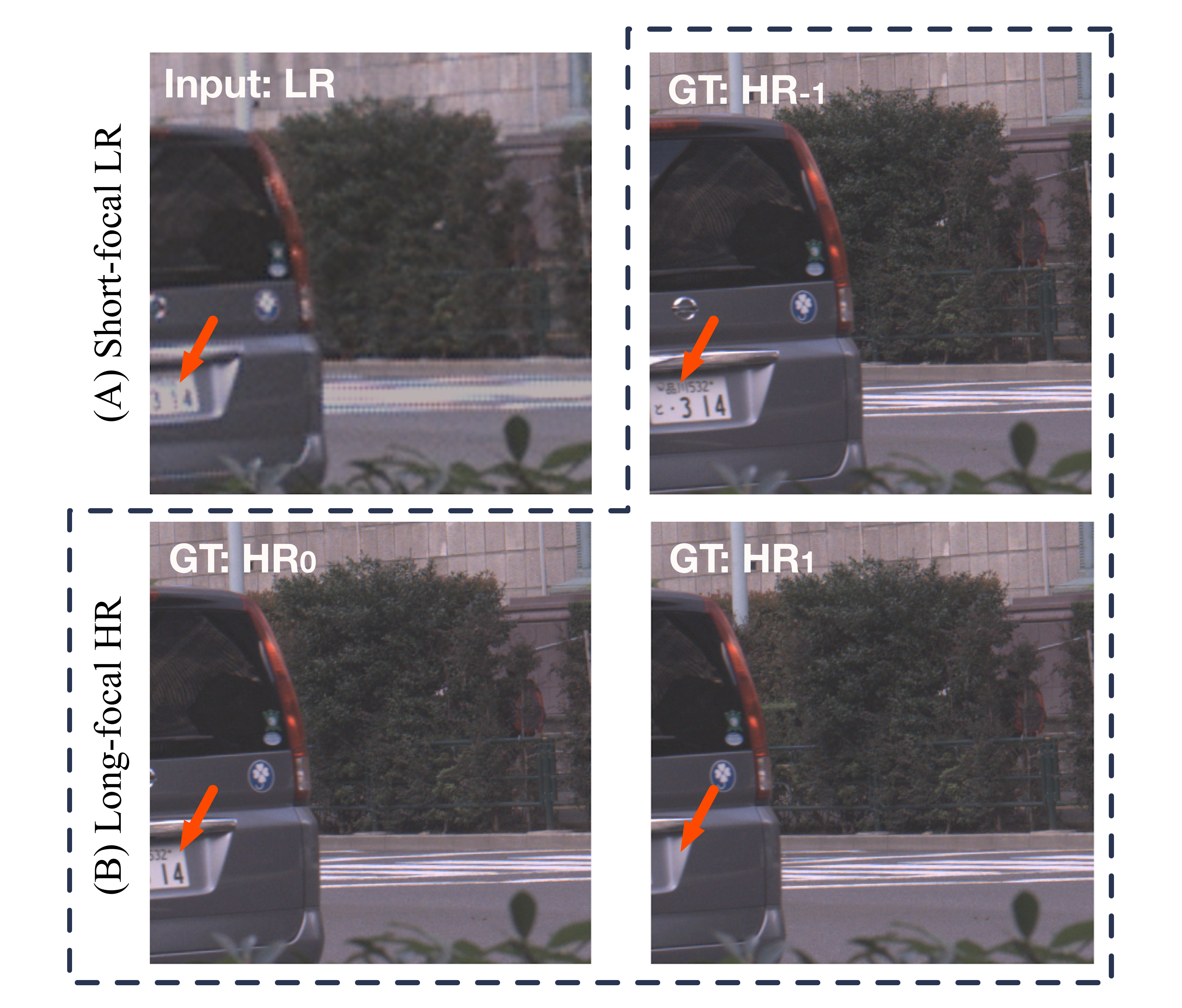}
\caption{Example pair (A and ground-truth: HR$_{0}$ in B) taken from corresponding training patches, the misalignment is unavoidable. Such issue motivates us to transit from single-frame to B consecutive multi-frame constraints.}
\label{fig:misalignment}
\end{figure}

In terms of geometric alignment, we first match the FoV of each paired LR and HR frame in VideoRAW based on the predefined focal lengths. Some examples are shown in Figure \ref{fig:scene_sample}. Since the videos taken at different focal lengths suffer from different lens distortions and perspective effects, as well as the subtle shift between light splitting paths, the misalignment is inevitable during data capturing. To address this issue, we employ homography transformation to warp the HR image based on the paired LR. Then, to match the size of LR frames based on the target zoom ratio, a scale offset is applied to HR frames. After that, we randomly crop consecutive frame patches from the paired videos for 4X SR training. Although obvious misalignment can be alleviated by the preprocessing step, as shown in Figure \ref{fig:misalignment} A and B (GT: HR$_{0}$), nontrivial misalignment between paired LR and HR imagery is still unavoidable. We usually observe 10-40 pixels shift in a processed pair depending on the scene geometry. We attribute this shift to various physical effects in optical zooming, such as perspective distortion, rather than the temporal synchronization, since the two cameras are rigorously synchronized within sub-microsecond accuracy. 

In terms of photometric alignment, including image brightness and color white balance, we capture the dark background and estimate white balance ratios between raw and RGB images for the blue and red channels. For all 16-bit raw images, we first subtract the black level, and then multiply the two ratios, so as to approximate the white balance of the RGB images. This correction allows to minimize the color and brightness difference between the pair, and helps to highlight the aforementioned discrepancies caused by the zooming effect. 

\section{Spatio-Temporal Coupling Loss}
\subsection{Framework}
\begin{figure*}[h!]
\centering
\includegraphics[width=\textwidth]{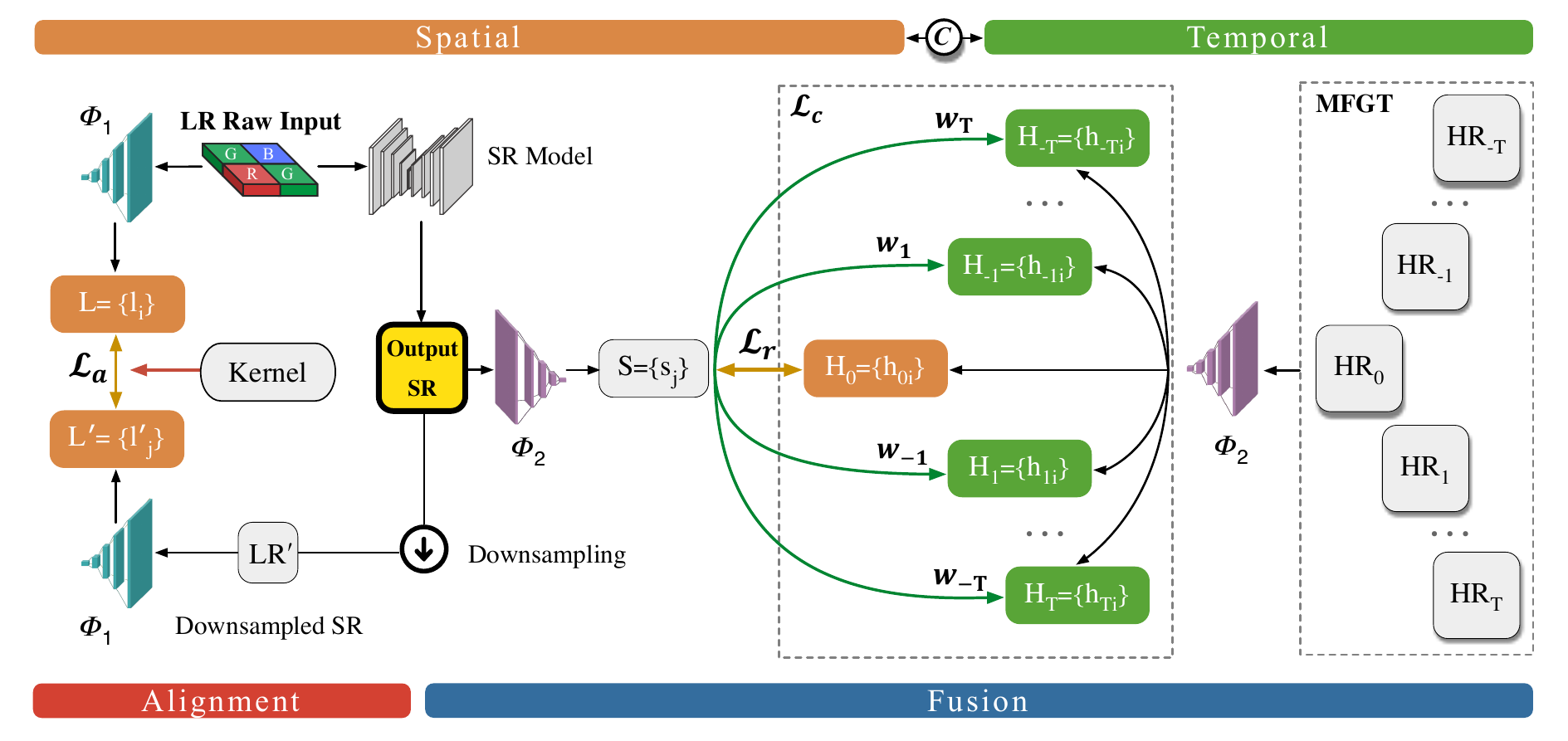}
\caption{Our framework for digital zoom quality enhancement based on consecutive multi-frame clips and spatio-temporal coupling.}
\label{fig:framework}
\end{figure*}

We propose a unified framework Spatio-Temporal Coupling Loss (STCL), as shown in Figure \ref{fig:framework}, which is extensible to different existing deep learning architectures for digital zoom quality enhancement. The challenge lies in the design of the constraint for spatio-temporal coupling via realistic video datasets when paired LR and HR$_{0}$ are misaligned, and establishing precise correspondences among LR and adjacent frames HR$_{t}$ is difficult. To obtain high-quality outputs, we solve the difficulties by (1) spatial alignment and fusion, and (2) temporal fusion and aggregation, both at the high-dimension feature level. 

Concretely, we design a spatial constraint by exploring the correlation among LR, HR$_{0}$, and expected SR. Considering in realistic scenarios when capturing a closer view of far-away subjects with greater details, the position relations among the features in zoom-in contents should rely on short-focal length LR, while the others such as edges, texture, and color are referenced based on long-focal length HR$_{0}$. Thus, the proposed spatial constraint consists of two components. The one aligns the feature position among SR and LR in a lower scale with a position constraint kernel, while the other one is responsible for fusing the features from the paired HR$_{0}$ frame into SR in higher scale. 

In terms of temporal constraint, we propose to compare the feature distribution of SR and adjacent frames rather than just comparing the appearance. Since different frames are not equally informative to the reconstruction, the weighted constraint is designed by considering the correlation between the features of HR$_{0}$ and adjacent frames. Thus, the temporal constraint is able to guide the feature extraction and aggregation from consecutive frames for the effective feature fusion and compensation. Finally, the spatio-temporal coupling can be achieved in the zoom task by the given realistic video pairs and the integration of spatial and temporal constraints.

\subsection{Loss Function}
Our objective function is formulated as

\begin{align} \label{update}
STCL = {Loss}_{s} + \lambda {Loss}_{t},
\end{align}
where $Loss_{s}$ and $Loss_{t}$ refer to spatial and temporal constraint, respectively. By effectively coupling spatial and temporal information from realistic video datasets, the STCL could achieve zoom quality enhancement. To the best of our knowledge, our approach proposed in this paper is the first attempt in this direction.

\noindent \textbf{Spatial constraint.} The core of $Loss_{s}$ is two loss terms: (1) The alignment loss, $Loss_{a}$, computed at low resolution, to drive the generated image to share the spatial structure of the LR. (2) The Contextual Loss (CX) \cite{mechrez2018contextual}, here defined as reference loss, ${Loss}_{r}$, is to make sure that the internal statistics of the generated image match those of the target $H_{0}$.

To align SR and LR, we first downsample SR into ${LR}'$ to match the size of LR in a lower scale. Instead of using pixel-to-pixel loss like $L_{1}$ and $L_{2}$, we align LR and ${LR}'$ at the feature level based on the features extracted by $[conv3\_2, conv4\_1]$ in pretrained VGG19 ($\Phi_{1}$) \cite{simonyan2014very}. Then we introduce spatial awareness into CX by Gaussian kernel to constraint the spatial distance between the two similar features. Our $Loss_{a}$ can be defined as

\begin{equation}\label{loss_a}
{Loss}_{a}(L,{L}') = \frac{1}{N}\sum_{i}^{N}\underset{j=1,...,M}{min}(\kappa \cdot {D}_{l_{i}, {l}'_{j}} ),
\end{equation}
where $L$ and ${L}'$ are the feature space in LR and ${LR}'$, respectively, and ${D}_{l_{i},{l}'_{j}}$ denominates the cosine distance between feature $l_{i}$ in $L$ and ${l}'_{j}$ in ${L}'$. The kernel $\kappa$ can be formulated as

\begin{equation}\label{kernel}
\kappa = exp(-\frac{{({D}'_{l_{i}, {l}'_{j}} - \mu })^{2}}{2\sigma ^{2}}),
\end{equation}
where ${D}'_{l_{i}, {l}'_{j}} = \left \| (x_{i}, y_{i}) - (x_{j}, y_{j}) \right \|_{2}$ denominates the spatial coordinate distance between feature $l_{i}$ and ${l}'_{j}$. Here, we select $\mu = 0$ and $\sigma = 2$. By adopting proposed $Loss_{a}$, similar feature pairs between LR and ${LR}'$ can be aligned spatially. 

Regarding ${Loss}_{r}$, since CX can be viewed as an approximation to KL divergence, and is designed for comparing images that are not spatially aligned \cite{mechrez2018maintaining}, we directly apply it to perform statistical constraint between feature distributions as 

\begin{equation}\label{distance_l}
{Loss}_{r}(H_{0},S) = \frac{1}{K_{0}}\sum_{i}^{K_{0}}\underset{j=1,...,G}{min}( {D}_{h_{0_{i}}, s_{j}} ),
\end{equation}
where $H_{0}$ and $S$ refer to the feature space generated by $[conv1\_2, conv2\_2, and conv3\_2]$ in VGG19 ($\Phi_{2}$). Thus, ${Loss}_{s} = {Loss}_{a}(L,{L}') + {Loss}_{r}(H_{0},S)$ can align feature from LR and fusion feature from $H_{0}$ spatially.

\noindent \textbf{Temporal constraint}. 
We further adopt CX to emphasize important features via temporal frames for information compensation and restoration. However, the adjacent frames are not equally beneficial to the reconstruction as $H_{0}$. To avoid the incorrectness brought by the adjacent frames which would decrease and corrupt the performance of SR, we define a correlation coefficient $w_{t}$ to weight each neighboring frame $H_{t}$. Here, the compensation loss for $S$ and $H_{t}$ is formulated as

\begin{align} \label{loss_c}
{Loss}_{c}(H_{t},S) = w_{t}\cdot CX(H_{t},S).
\end{align}

Then, the temporal loss is defined by aggregating all the compensation losses as

\begin{align} \label{loss_t}
{Loss}_{t} = \sum_{t=-T}^{T} \cdot {Loss}_{c} (H_{t},S), t \neq 0.
\end{align}

In this paper, since our current dataset (15fps) mainly covers city views with pedestrians and moving vehicles under the speed of 45km/h, we choose T = 1 (3-frame clip) and $w_{\pm}$ = 0.1.  

\section{Experimental Setup}
% 10.28
16-bit LR raw and 8-bit HR RGB videos are adopted to train a 4X SR model. We first randomly choose 80 clips from different video pairs in VideoRAW, with 4000 image pairs, for training, validation, and testing. The ratio of them is about 45:10:45. Then, we randomly crop 160$\times$160 and 640$\times$640 consecutive patches from LR and HR clips as input for training. Here, 16 layer ResNet \cite{he2016deep} based SRResNet \cite{ledig2017photo} without batch normalization \cite{wang2018esrgan} is adopted for SR architecture. We select a batch size of one, thus in our spatio-temporal model, one LR Bayer mosaic would pair three consecutive HR RGB ground truth frames for each iteration. We implement the proposed network based on TensorFlow 1.9 and train it with NVIDIA Tesla V100. The proposed model is trained for 200,000 iterations with 100 validations performed by every 1,000 iterations. In our experiment, parameters are optimized by the Adam optimizer \cite{kingma2014adam} using initial learning rate = $1e^{-4}$, $\beta_{1}$ = 0.9, $\beta_{2}$ = 0.999, and $\epsilon$ = $1e^{-8}$.

Given the existing video SR models are not designed for realistic dataset with misalignment issues, in the baseline, we propose to investigate the feasibility of achieving spatio-temporal coupling in the perspective of loss function. We first compare the proposed spatio-temporal coupling approach to the loss functions which rely on distribution constraint, weak-spatial constraint, and spatial constraint, respectively. Here we define pixel-wise methods as weak-spatial in misalignment condition. Then we conduct the ablation study on our model variants on spatial compensation. After that, we integrate our framework into other deep learning architectures for generalization testing. Finally, we investigate the extensibility of our approach into more challenging video zoom tasks by perceptual experiments. All comparisons are conducted on the random selected three scenes, and each contains 4 clips with 200 frames.

% Given the existing video SR models are not designed for realistic dataset with misalignment issues, in the baseline, we propose to investigate the feasibility of achieving spatio-temporal coupling in the perspective of loss function.
% To demonstrate the effectiveness of the proposed spatio-temporal coupling approach, we first compare it to the loss functions which rely on distribution constraint, weak-spatial constraint, and spatial constraint, respectively. Here we define pixel-wise methods as weak-spatial in misalignment condition. Then we conduct the ablation study on our model variants on spatial compensation. After that, we integrate our framework into other deep learning architectures for generalization testing. Finally, we investigate the extensibility of our approach into more challenging video zoom tasks by perceptual experiments. All comparisons are conducted on the random selected three scenes, and each contains 4 clips with 200 frames.

\subsection{Baselines}
For comparison, we choose a few representative loss functions used for SR methods based on spatial and temporal concern: CX \cite{mechrez2018contextual}, which performed by comparing statistic feature distributions without considering both spatial and temporal correlations; $L_{2}$ \cite{ledig2017photo}, the most widely used pixel-wise spatial constraint applied in many state-of-the-art SR approaches; CoBi \cite{zhang2019zoom}, an effective loss used in realistic SR for spatial constraint. All the baselines are integrated into SRResNet architecture for re-training on the proposed training dataset.

\section{Results and Discussions}

\begin{figure*}[ht]
\centering
\includegraphics[width=\textwidth]{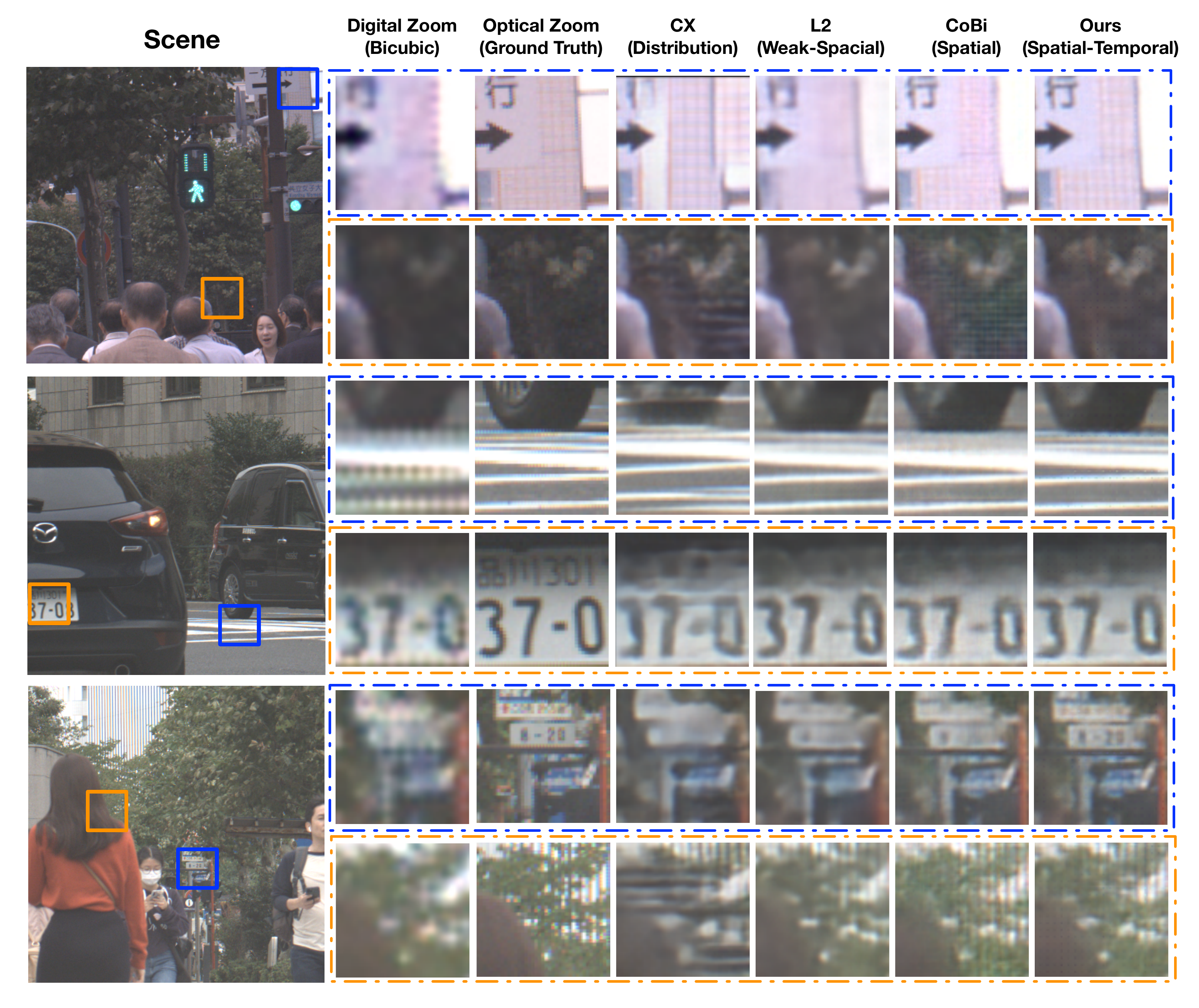}
\caption{Our zoom results show better perceptual performance against baseline methods in different scenes.}

\label{fig:qual}
\end{figure*}

\subsection{Quantitative Evaluation}
To evaluate our method as well as the baselines, evaluation metrics, including pixel-based PSNR, structure-based SSIM, and learning-based LPIPS, are adopted. Unlike the case of PSNR and SSIM, the lower score of LPIPS indicates better image quality.

The relative performances of different methods over testing data are listed in Table \ref{tab:quan}. In general, the proposed method outperforms others in terms of most evaluation metrics and scenes, while CX, enforcing constraints on feature distribution only, gets the worst in learning based methods. Specifically, both our method and CoBi are better than $L_{2}$ in PSRN. The $L_{2}$ loss function usually achieves particularly high PSNR in SR than others due to the pixel-to-pixel mapping via MSE. However, in realistic case where LR and HR are misaligned, such pixel-wise mapping will bring incorrectness and noise to learning, which yields the lower PSNR performance. It indicates that the pixel-wise loss cannot perform effective mapping between LR and misaligned HR. In perspective of LPIPS, since the pixel-wise optimization often lacks high frequency content, it results in perceptually unsatisfying results with overly smooth textures. Our model and CoBi still perform much better than $L_{2}$ in all scenes. Besides, by adopting spatio-temporal coupling, our model shows better performance than CoBi in all cases. Such results verify the effectiveness of introducing temporal components in zoom quality enhancement.

\begin{table*}[h]
\small
\centering
\caption{Performance comparison on digital zoom tasks under different scenes. Metric with '$\uparrow$' means the higher the better image quality, while '$\downarrow$' means the opposite.}
\begin{tabular}{cccccccccccc}
\toprule
\textbf{Scene}     & \multicolumn{3}{c}{\#1}                              &  & \multicolumn{3}{c}{\#2}                              &  & \multicolumn{3}{c}{\#3}                             \\
\midrule
\textbf{Method}  & \textbf{PSNR $\uparrow$}             & \textbf{SSIM $\uparrow$}            & \textbf{LPIPS$\downarrow$}           &  & \textbf{PSNR$\uparrow$}             & \textbf{SSIM$\uparrow$}            & \textbf{LPIPS$\downarrow$}           &  & \textbf{PSNR$\uparrow$}             & \textbf{SSIM$\uparrow$}           & \textbf{LPIPS$\downarrow$}           \\
\midrule
Bicubic & 12.5652          & 0.4584          & 0.6525          &  & 11.5330           & 0.4928          & 0.5081          &  & 11.6146          & 0.4637         & 0.5888          \\
CX \cite{mechrez2018contextual}     & 24.4284          & 0.6652          & 0.3900            &  & 24.2536          & 0.7503          & 0.3582          &  & 25.0389          & 0.7134         & 0.3355          \\
$L_{2}$ \cite{ledig2017photo}    & 29.4006          & 0.8034          & 0.3456          &  & 26.8314          & \textbf{0.8419} & 0.3011          &  & 26.5033          & 0.7914         & 0.3208          \\
CoBi \cite{zhang2019zoom}   & 29.4692          & 0.8131          & 0.2336          &  & 27.9387          & 0.8272          & 0.2207          &  & 27.1417          & 0.7759         & 0.2442          \\
Ours    & \textbf{30.2093} & \textbf{0.8216} & \textbf{0.2213} &  & \textbf{27.9551} & 0.8311          & \textbf{0.2114} &  & \textbf{27.4081} & \textbf{0.7930} & \textbf{0.2391} \\
\bottomrule
\end{tabular}
\label{tab:quan}
\end{table*}

\subsection{Qualitative Evaluation}
Qualitative comparison of our model against baselines in all three scenes is shown in Figure \ref{fig:qual}.  Within these scenes, moving vehicles and pedestrians are presented. Since the direct Bicubic upsampling from LR (the $2^{nd}$ column) brings very blurry outlook, and the CX (the $4^{th}$ column) leads strong artifacts that caused by the inappropriate feature matching, we mainly focus on the comparison between 'weak-spatial' $L_{2}$, 'spatial' CoBi, and our 'spatiao-temporal'.

In scene 1, we focus on the comparison of character (the $1^{st}$ row) and vegetation (the $2^{nd}$ row). Due to the weak-spatial mapping, $L_{2}$ results in very blurry for the character. Although CoBi can generate clear character as ours, for the high frequency texture back on the wall, our method achieves sharper edges and finer textures. As for vegetation, our method yields the most consistent visual result without any artifacts. In scene 2, it appears that our method can super-resolve zebra crossing (the $3^{th}$ row) with higher quality, while CoBi is too pale with limited contrast. Regarding the number plate on the moving vehicle (the $4^{th}$ row), although all the results generated by baselines and ours are not very clear, which we think is caused by high noise and signal loss in original LR image, our method still generates better results. In scene 3, our method results in very clear appearance for distant guideboard (the $5^{th}$ row) and wall (the $6^{th}$ row), while $L_{2}$ is too blurry to see the details and CoBi yields additional unnatural 'stripe' artifacts on the vegetation. All of the above qualitative results demonstrate the effectiveness of our method on the realistic SR in different scenarios.

\subsection{Ablation Analysis}
To further investigate the effectiveness of the proposed spatio-temporal coupling method, we conduct an ablation study using two variants: one with temporal compensation provided by multi-frame videos and the other without. The relative quantitative comparison is shown in Table \ref{tab:ablation}.

\begin{table}[ht]
\caption{Performances of variants with or without temporal compensation. 'T' refers to temporal compensation.}
\centering
\small
\begin{tabular}{c cc c c c }
\toprule
 \multicolumn{1}{c}{\textbf{Scene}} &  \multicolumn{2}{c}{\textbf{Method}} & \textbf{PSNR$\uparrow$}& \textbf{SSIM$\uparrow$} & \textbf{LPIPS$\downarrow$} \\
 
 \midrule
 \multirow{2}{*}{\#1} 
 &  \multicolumn{2}{c}{Ours(-T)}& 29.7731 &0.7865	&0.2254  \\
 &  \multicolumn{2}{c}{Ours(+T)}& \textbf{30.2093}	&\textbf{0.8216}	&\textbf{0.2213}  \\

\midrule
 \multirow{2}{*}{\#2} 
 &  \multicolumn{2}{c}{Ours(-T)}& 27.7265	& 0.8105	& 0.2123  \\
 &  \multicolumn{2}{c}{Ours(+T)}& \textbf{27.9551} & \textbf{0.8311}	& \textbf{0.2114}  \\

\midrule
 \multirow{2}{*}{\#3} 
 &  \multicolumn{2}{c}{Ours(-T)}& 27.0743	&0.7462	&\textbf{0.2309}  \\
 &  \multicolumn{2}{c}{Ours(+T)}& \textbf{27.4081}&	\textbf{0.7930}	&0.2391  \\

\bottomrule
\end{tabular}
\label{tab:ablation}
\end{table}

% \midrule
%  \multirow{2}{*}{\#average} 
%  &  \multicolumn{2}{c}{Ours(-T)}& 28.1793	&0.7811	&\textbf{0.2229}  \\
%  &  \multicolumn{2}{c}{Ours(+T)}& \textbf{28.5242}&	\textbf{0.8152}	&0.2239  \\

It reveals that with the help of spatio-temporal coupling, the reconstruction quality on multiple metrics can be improved, which consolidates the value of temporal compensation. In Particular, for SSIM, it leads to about 4.4\% (0.8152 vs. 0.7811) improvement on average.

\subsection{Generalization Ability}
\begin{table}[ht]
\small
\caption{Generalization ability analysis using existing deep learning architectures.}
\centering
\resizebox{.9\columnwidth}{!}{\begin{tabular}{ccccc }
%\begin{tabular}
\toprule
  \multicolumn{1}{c}{\textbf{Archi.}} &  \textbf{Method} & \textbf{PSNR$\uparrow$}& \textbf{SSIM$\uparrow$} & \textbf{LPIPS$\downarrow$} \\
 \midrule
 \multirow{4}{*}{EDSR \cite{lim2017enhanced}} 
 &CX \cite{mechrez2018contextual} & 25.7379 &0.7828 & 0.3679  \\
&Ori & 26.6179 &0.7798 & 0.3724  \\
 &CoBi \cite{zhang2019zoom} & 27.7801 & 0.7918 & 0.3405  \\
 &Ours & \textbf{27.8733} & \textbf{0.7953} & \textbf{0.3248}  \\
  \midrule
 \multirow{4}{*}{DCSCN \cite{yamanaka2017fast}} 
 &CX \cite{mechrez2018contextual} & 25.2992 &0.8035 & 0.3488  \\
&Ori & 26.5322 &\textbf{0.8318} & 0.3471  \\
 &CoBi \cite{zhang2019zoom} & 27.5385 & 0.8226 & 0.3130  \\
 &Ours & \textbf{27.8891} & 0.8228 & \textbf{0.2946}  \\
  \midrule
 \multirow{4}{*}{FEQE \cite{vu2018fast}} 
 &CX \cite{mechrez2018contextual} & 25.7089 &0.7937 & 0.3374  \\
&Ori & 27.0253 &0.8287 & 0.3623  \\
 &CoBi \cite{zhang2019zoom} & 27.7957 & 0.8360 & 0.3134  \\
 &Ours & \textbf{27.9870} & \textbf{0.8376} & \textbf{0.2913}  \\
\bottomrule
\end{tabular}}
\label{tab:gener}
\end{table}

To evaluate the generalizability of the proposed method, we further integrate our framework into more existing deep learning architectures. In this paper, EDSR \cite{lim2017enhanced}, DCSCN \cite{yamanaka2017fast}, and FEQE \cite{vu2018fast} are adopted. For comparison, the 'Ori' refers to the weak-spatial loss function applied in the original paper. The comparison results are generated from the average performance of three scenes, as shown in Table \ref{tab:gener}, which reveals the generalization of the proposed method.

\subsection{Perceptual Experiments for Video Zoom}

\begin{table}[ht]
\caption{Perceptual experiments show that our results are significantly preferred on video zoom tasks.}
\small
\centering
\resizebox{\columnwidth}{!}{\begin{tabular}{c ccccc }
\toprule
  \multirow{2}{*}{\textbf{Scene}} &  \multicolumn{4}{c}{\textbf{Preference Rate}} \\
 \cmidrule(lr){2-6} &  CX \cite{mechrez2018contextual} & $L_{2}$ \cite{ledig2017photo}& CoBi \cite{zhang2019zoom} & Ours & No preference\\
\midrule
 \#1 & 3.33\%	&6.67\%  	&15.00\%	&\textbf{53.33\%}	&21.67\%  \\
 \#2 & 1.67\%	&23.33\%	&8.33\%	    &\textbf{55.00\%}	&11.67\%  \\
 \#3 & 1.67\%	&18.33\%	&15.00\%	&\textbf{50.00\%}	&15.00\% \\
 \bottomrule
\end{tabular}}
\label{tab:perceptual}
\end{table}

Moreover, to demonstrate the proposed method has a favorable capability in terms of video zoom, we evaluate the perceptual quality of the generated video through blind testing. In each inquiry, we present the participants with three videos (200 frames per video) taken from different scenes. At every frame, ground truth image and corresponding images generated from baseline models (CX, $L_{2}$, CoBi) and ours are organized side by side. The location information of both 'Ours' and the 'Baseline' are not provided. The participants are asked to pick up the one that is more close to the ground truth video. In the experiment, the responses from 60 valid participants are collected and listed in Table \ref{tab:perceptual}. Since some occasional but noticeable artifacts in CoBi might severely influence subjective evaluation, especially in \#2, videos generated by ours achieve a significantly higher preference rate under blind pairwise human judgment.

\section{Conclusion}
This paper investigated the effectiveness of spatial and temporal coupling for digital zoom quality enhancement. To enable training with spatio-temporal information, we collect a new dataset that contains realistic LR\&HR video pairs, and introduce a novel loss framework for spatio-temporal constraint. The experimental results demonstrated the potential and capability of the proposed method in solving realistic SR problems. 

% \vfill

% \newpage

{\small
\bibliographystyle{ieee_fullname}
\bibliography{egpaper_final}
}

\end{document}